\documentclass[letterpaper, 10 pt, conference]{ieeeconf}

\IEEEoverridecommandlockouts
\usepackage{stfloats}
\usepackage[ruled,vlined,linesnumbered]{algorithm2e}
\SetAlgoSkip{}%
\makeatletter%
\def\@algocf@post@ruled{\kern\interspacealgoruled\hrule width\algocf@ruledwidth height\algoheightrule\relax\kern-9pt}
\makeatother
\SetKwInOut{Input}{Input}
\SetKwInOut{Output}{Output}
\usepackage{amsmath,amssymb,mathtools}

\usepackage{float}
\usepackage{graphicx}
\usepackage{graphics}
\usepackage{epsfig}
\usepackage{mathptmx}
\usepackage{times}
\usepackage{hyperref}
\newcommand{\videolink}{Videos: \url{https://hybrid-imitation.github.io/}.}
\usepackage{multirow}
\usepackage[table]{xcolor}
\usepackage{comment}
\usepackage{pifont}

\usepackage{bm}
\usepackage{footmisc}
\usepackage{balance}
\usepackage{cite}
\usepackage{booktabs}

\newcommand{\SqDemos}{165}
\newcommand{\MgDemos}{169}
\newcommand{\SqUnassisted}{0.372}
\newcommand{\SqTap}{0.531}
\newcommand{\SqTapbase}{0.297}
\newcommand{\SqP}{2.1\times10^{-18}}
\newcommand{\SqPTapbaseVsUn}{1.7\times10^{-5}}
\newcommand{\SqPTapbaseVsTap}{7.1\times10^{-39}}
\newcommand{\SqStageUn}{0.207}
\newcommand{\SqStageTap}{0.225}
\newcommand{\SqPStage}{0.27}
\newcommand{\SqPGiven}{2\times10^{-26}}
\newcommand{\SqFiredNoLift}{163}
\newcommand{\SqWrongPerch}{219}
\newcommand{\SqGoodFires}{929}
\newcommand{\SqGoodFireSucc}{0.784}
\newcommand{\SqEpisodes}{1500}
\newcommand{\SqGivenUn}{0.469}
\newcommand{\SqGivenTap}{0.685}
\newcommand{\MgUnassisted}{0.203}
\newcommand{\MgTap}{0.323}
\newcommand{\MgTapbase}{0.243}
\newcommand{\MgP}{9.3\times10^{-14}}
\newcommand{\MgPTapbaseVsUn}{8.5\times10^{-3}}
\newcommand{\MgPTapbaseVsTap}{1.7\times10^{-6}}

\newcommand{\MgGivenUn}{0.412}
\newcommand{\MgGivenTap}{0.692}
\newcommand{\MgPGiven}{1.2\times10^{-26}}
\newcommand{\MgEpisodes}{1500}
\newcommand{\MgStageUn}{0.509}
\newcommand{\MgStageTap}{0.534}
\newcommand{\MgPStage}{0.18}
\newcommand{\MgLatchPressed}{1329}
\newcommand{\MgLatchErrCm}{1.2}
\newcommand{\MgLatchErrYaw}{2.1}
\newcommand{\MgLatchErrCmHi}{3.0}
\newcommand{\MgLatchErrYawHi}{8.8}
\newcommand{\MgCarryPressed}{681}
\newcommand{\MgCarryHeld}{575}
\newcommand{\MgCarryHeldSucc}{0.824}
\newcommand{\MgCarryEmpty}{106}
\newcommand{\SqMisfire}{20}
\newcommand{\MgMisfire}{7}

\newcommand{\SqPctRoutine}{17}

\newcommand{\SqPairedWL}{163/72, 149/73, 147/75}
\newcommand{\SqPairedP}{2.8\times10^{-9}, 3.7\times10^{-7}, 1.5\times10^{-6}}
\newcommand{\MgPairedWL}{120/58, 114/49, 108/55}
\newcommand{\MgPairedP}{3.9\times10^{-6}, 3.8\times10^{-7}, 4\times10^{-5}}

\newcommand{\SqBcUn}{0.085}
\newcommand{\SqBcTap}{0.250}
\newcommand{\SqBcP}{2\times10^{-34}}
\newcommand{\SqBcTransformerUn}{0.263}
\newcommand{\SqBcTransformerTap}{0.329}
\newcommand{\SqBcTransformerP}{7\times10^{-5}}

\newcommand{\MgBcUn}{0.025}
\newcommand{\MgBcTap}{0.074}
\newcommand{\MgBcP}{3\times10^{-10}}
\newcommand{\MgBcTransformerUn}{0.106}
\newcommand{\MgBcTransformerTap}{0.153}
\newcommand{\MgBcTransformerP}{2\times10^{-4}}

\newcommand{\MgNogate}{0.317}
\newcommand{\MgNogateP}{0.75}
\newcommand{\MgGateGain}{0.006}

\newcommand{\MgGateBlocked}{466}

\newcommand{\tap}{TAP}
\newcommand{\taps}{TAPs}
\newcommand{\method}{Hybrid Imitation Learning}

\title{\LARGE \bf
Hybrid Imitation Learning: Teleoperation Augmentation Primitives that Policies Learn to Trigger
}

\author{Jonne Van Haastregt*$^1$, Bastian Orthmann*$^1$,  Michael C. Welle*$^{1,2}$, Yuchong Zhang$^2$, Danica Kragic$^2$%
\thanks{*These authors contributed equally.}%
\thanks{1-INCAR Robotics AB, Sweden }
\thanks{2-KTH Royal Institute of Technology, Sweden}
}

\begin{document}
\maketitle
\thispagestyle{empty}
\pagestyle{empty}

\begin{abstract}
What an operator can demonstrate bounds what imitation learning can learn. Teleoperation interfaces map the human body to the robot, so motions that are hard for a human, such as holding an exact orientation, returning to the same viewpoint, or turning a wrist joint several full revolutions, are difficult to demonstrate on most teleoperation interfaces, even when they are trivial for the robot. We introduce \emph{Teleoperation Augmentation Primitives} (\taps{}): axis locks, perching waypoints, pose anchors, and embodiment-specific routines that the operator triggers during a demonstration by speech, an AR menu, or simply with a controller button.
\taps{} are themselves recorded in the demonstration and can therefore also be learned and invoked by the policy itself.
In simulation, where benchmarks provide human demonstrations, we show that augmenting them with \taps{} after the fact yields better policies on a wrist-camera-only peg insertion ($\SqUnassisted$ to $\SqTap$ success) and on a mug-cleanup task with a learned ``remember this pose'' anchor ($\MgUnassisted$ to $\MgTap$). On a real robot, three tasks show the same pattern: primitives that help the operator do not hurt the policy, and a policy triggering an unscrewing routine succeeds where a plain policy cannot resolve the multi-turn motion from images alone ($67\%$ vs.\ $38\%$ success). We close with an industrial proof-of-concept case leveraging this idea. \videolink{}
\end{abstract}

\section{Introduction}

Imitation learning (IL) has enabled advanced robotic manipulation tasks such as spreading sauce on a pizza~\cite{chi2023diffusion}, opening a capped bottle~\cite{ingelhag2024robotic}, and installing a bike rotor~\cite{barreiros2025careful}. Performance depends heavily on demonstration quality~\cite{barreiros2025careful,ingelhag2025real,mandlekar2021matters}, hence on the teleoperation interface and the operator's skill; an ever-growing body of work therefore addresses different interfaces: 6D mice~\cite{chi2023diffusion}, VR/AR controllers~\cite{welle2024quest2ros,van2024puppeteer}, puppeteering hardware~\cite{fu2024mobile,fang2024airexo,jiang2025behavior}, and whole-body setups~\cite{he2024omnih2o,ze2025twist}.
However, a human struggles to hold an exact orientation while translating~\cite{zhai1998quantifying}, cannot return to the exact same viewpoint twice, and cannot easily turn a wrist through several full revolutions; a robot does all three trivially, and a joint with no human counterpart~\cite{BostonDynamics_AtlasElectric_2024} cannot be demonstrated in a straightforward way. Since demonstration quality bounds what these methods can learn, motions made awkward by the interface tend to enter the data in a degraded form, and the resulting policy may inherit that degradation.
\begin{figure}[t]
    \centering
    \includegraphics[width=0.85\linewidth]{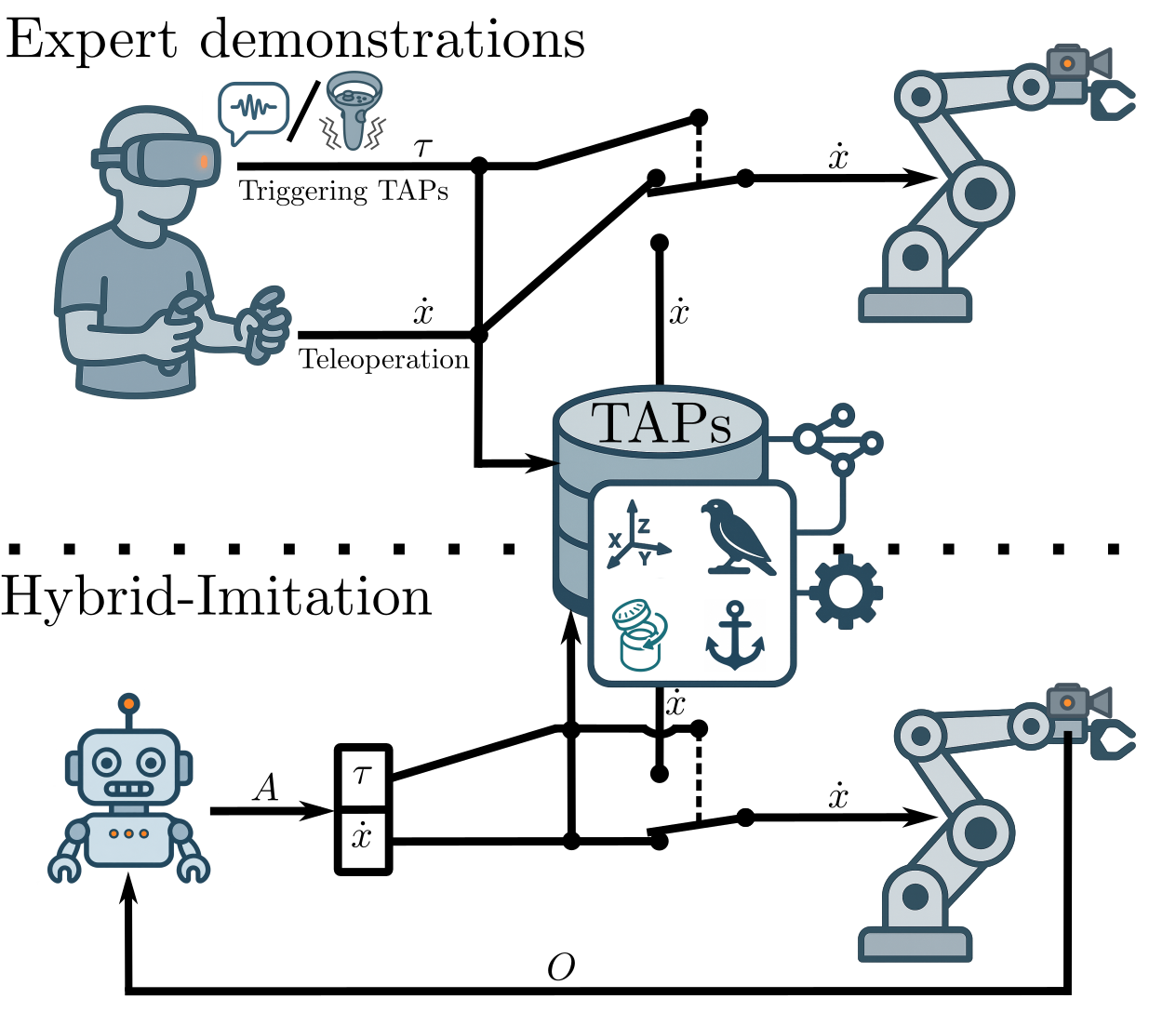}
    \caption{Overview. During teleoperation the operator triggers a Teleoperation Augmentation Primitive (\tap{}) by speech or (AR) controller input; the press and the primitive's motion are recorded with the demonstration. The policy is trained with a discrete primitive channel next to its continuous action and learns to trigger the same primitives at inference time.}
    \label{fig:hybrid_diffusion}
    \vspace{-\baselineskip}
\end{figure}
We propose to let the operator abstract parts of a demonstration and leave the realisation to the robot. \emph{Teleoperation Augmentation Primitives} (\taps{}) are primitives the operator triggers mid-demonstration through speech, an AR menu, or a controller button. We introduce four different kinds. \emph{Axis locks} fix selected translational or rotational axes in the tool or base frame so that a reached orientation is held regardless of hand tremor. \emph{Perching waypoints} move the end-effector to a pre-set pose, which brings a scene back into an eye-in-hand camera and effectively splits a long task into shorter segments with common waypoints. \emph{Anchors} record the current pose so that a later primitive can be expressed relative to it (``remember this pose''). \emph{Routines} execute an embodiment-specific motion, such as multi-turn unscrewing, that the human cannot perform through the interface. \taps{} are cheap to author; they are triggered by the operator during demonstration collection and become part of the policy later on.
Furthermore, we add a discrete primitive channel to the action representation of imitation learning policies; the policy predicts it alongside the continuous action, and a rising edge on the channel hands control to the primitive until it finishes.
In simulation we instantiate the channel on three policy classes spanning three architectures: Diffusion Policy~\cite{chi2023diffusion}, a feed-forward behavioural-cloning MLP (BC)~\cite{mandlekar2021matters}, and a causal-Transformer behavioural-cloning policy (BC-Transformer)~\cite{mandlekar2021matters}, each trained on the same demonstrations with the same encoder, budget, seeds and evaluation (Sec.~\ref{sec:sim_baselines}). On the real robot we use Diffusion Policy alone, and refer to that instantiation as \emph{Hybrid-Diffusion}.

One issue with evaluating our \tap{} setup against unaided demonstrations is that it is hard to obtain unbiased demonstrations from the same expert operator who knows and uses a \tap{} system. We therefore make use of simulated settings, where unbiased and widely used demonstrations are available. The existing unbiased demonstrations are relabelled by replaying them with a \tap{} inserted at the operator's natural hand-over point, and the same demonstrations, with and without the primitive, train the same policy under the same seeds.
Two tasks are used: a wrist-camera-only peg insertion with a perching waypoint, and a mug-cleanup task where the policy learns to press an anchor before opening the drawer and a carry routine to that anchor after grasping the mug. In both, \tap{}-collected demonstrations yield a better policy for the deployed imitation learning methods.
We also compare against policies that replace \taps{} by ordinary actions. The trajectories in the data are then essentially identical, but no trigger is learned, which isolates what the learned trigger itself contributes. On the real robot the difference is clearest where a routine is visually hard to distinguish from the motion around it: on unscrewing, the policy without the trigger lifts the cap before it is unscrewed, since a closed cap and a cap turned once look alike, and succeeds in $8$ of $21$ rollouts against $14$ for the triggering policy. Across all three real tasks no \tap{}-triggering policy performs worse than its counterpart, so a primitive that already helps the operator costs the policy nothing and can take over a motion the policy cannot resolve. The contributions of the work are:
\begin{enumerate}
    \item \taps{}: a formalism for Teleoperation Augmentation Primitives with four types (Sec.~\ref{sec:method}).
    \item A primitive channel and execution loop that lets an action-chunking policy hand control to a primitive and take it back, and that carries across all the policy classes tested here (Sec.~\ref{sec:channel}).
    \item A robust controlled simulation study showing that \tap{} demonstrations can lead to significant improvements across policy classes (Sec.~\ref{sec:sim}).
    \item Real-robot validation on three tasks (Sec.~\ref{sec:real}) and an industrial proof-of-concept of policy-triggered routines on a long-horizon assembly (Sec.~\ref{sec:deploy}).
\end{enumerate}

\section{Related Work}

\noindent\textbf{Hybrid action spaces in imitation learning.} Several works shorten the effective horizon of visuomotor IL by mixing sparse and dense control. HYDRA~\cite{belkhale2023hydra} alternates between predicted waypoints and per-timestep actions and relabels demonstrations for consistency; AWE~\cite{shi2023waypoint} extracts waypoints automatically from demonstrations; SPHINX~\cite{sundaresan2024s} anchors waypoints to salient 3D points and switches to dense wrist-image control near contact; HOMER~\cite{sundaresan2025homer} extends the idea to mobile manipulation. In all of these the sparse mode is a \emph{predicted pose} and the dense mode is the policy. Our primitives are richer than a waypoint (a multi-turn routine, an axis lock, a pose anchor) and, more importantly, they are chosen and triggered by the \emph{operator} during the demonstration, so the switching labels are part of the data rather than the product of a post-hoc segmentation.

\noindent\textbf{Primitive-augmented policies.} MAPLE~\cite{nasiriany2022maple} and PRIME~\cite{gao2024prime} give a learner a library of parameterised behaviour primitives (grasp, push, reach) and learn to sequence them; SayCan~\cite{ahn2022can} and Code-as-Policies~\cite{liang2023code} let a language model select skills. Diffusion-based policies have likewise been combined with skill libraries~\cite{liang2023skilldiffuser,xian2023chaineddiffuser}. These approaches sequence primitives \emph{instead of} low-level control or learn a high-level policy over them. We keep the dense visuomotor policy as the default and insert a primitive only where the operator did; the policy's only extra output is a discrete channel that is trained by supervised learning from the recorded presses, and the primitive's parameters can come from the policy's own anchor presses.

\noindent\textbf{Assisted teleoperation and virtual fixtures.} Constraining the operator's motion to make demonstrations easier is an old idea: virtual fixtures~\cite{rosenberg1993virtual} and shared-control blending~\cite{dragan2013policy,xu2022shared} guide or partially automate the commanded motion. Our axis lock is a virtual fixture and our perching waypoint is an automated sub-motion. The difference is what happens after the demonstration: the fixture is recorded as an explicit event that the learned policy can reproduce, whereas assisted-teleoperation systems discard the assistance once the data is collected, and the policy must then learn the assisted motion as if it had been demonstrated by hand.

\noindent\textbf{Interactive and intervention-based imitation learning.} DAgger-style methods~\cite{ross2011reduction,kelly2019hgdagger,menda2019ensembledagger} and intervention-based systems~\cite{mandlekar2020human,spencer2020learning,liu2022robot,ingelhag2025real} let a human correct the policy during rollouts and learn from the corrections. \taps{} are orthogonal: they act on the demonstration itself, and they add capabilities to the policy (a routine it cannot execute from vision alone) rather than data in the policy's own state distribution. The two combine naturally, since an operator can trigger a \tap{} during an intervention as well.

\noindent\textbf{Teleoperation interfaces.} The interface builds on VR/AR teleoperation~\cite{galarza2023virtual,moletta2023virtual,welle2024quest2ros,arevalo2021assisting} and specifically on the puppeteer framework~\cite{van2024puppeteer} and its voice- and gesture-driven extensions~\cite{zhang2025llm,zhang2025multimodal}, whose base usability has been evaluated in those works. We add the \tap{} triggering layer (speech, AR menu, haptic-confirmed buttons) on top; the usability of the \tap{} layer itself is not the subject of this paper.

\section{Methodology} \label{sec:method}

\subsection{Teleoperation Augmentation Primitives}

A \tap{} is a tuple $(\tau, \rho, \theta)$: a \emph{trigger} $\tau$ (the operator's press during demonstration; the policy's channel at execution), a \emph{routine} $\rho$ that takes over the robot from the trigger until it terminates, and \emph{parameters} $\theta$. Parameters come from: static configuration (a waypoint pose, the number of turns of a routine); \emph{anchors}, poses recorded by the operator before the demonstration (a perching waypoint) or during it (an anchor press); and the robot's proprioception at the trigger.

\begin{figure*}[t]
    \centering
    \includegraphics[width=0.95\linewidth]{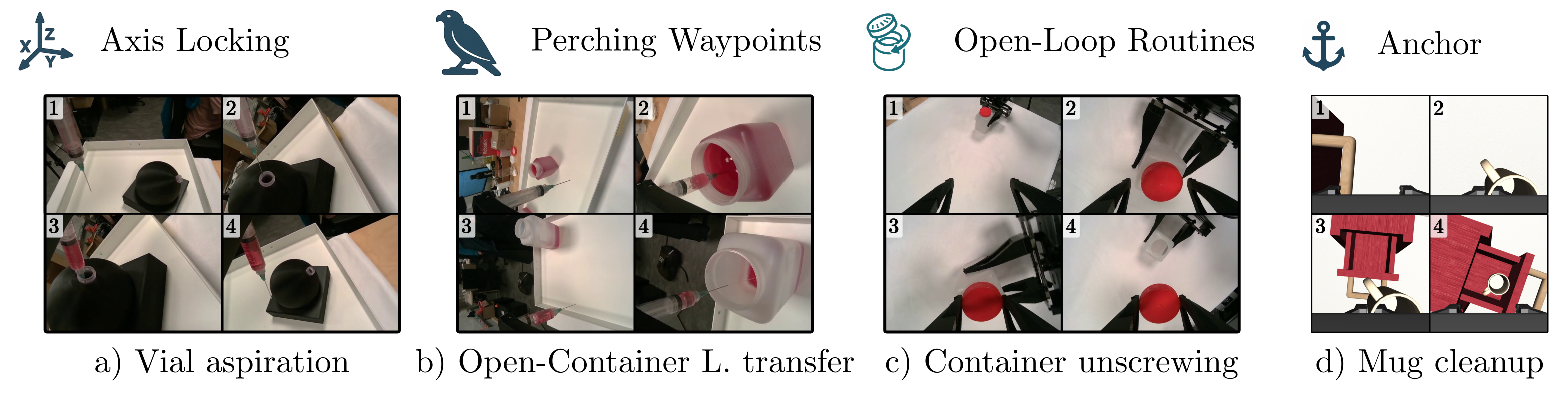}
    \vspace{-\baselineskip}
    \caption{\tap{} types: (a)~Vial aspiration: operator locks the rotational axes once the needle is aligned. (b)~Open-container liquid transfer: perching waypoints bring the source and the target container into the eye-in-hand view. (c)~Container unscrewing: an unscrewing routine grasps, unscrews and lifts the cap. (d)~MugCleanup (simulated): an anchor press stores the drawer frame while the open gripper sits behind the handle, the last step before the drawer moves; after the mug is grasped, a carry routine transports it into that frame and shuts the drawer.}
    \label{fig:taptypes}
    \vspace{-\baselineskip}
\end{figure*}

\noindent\textbf{Axis locks.} Mimicking the operator's 6D hand pose is the common teleoperation mapping, and adding rotation to translation is known to be hard for humans~\cite{zhai1998quantifying,tugal2025operator,wang2016multi,zhang2024usability}. An axis lock fixes one or more of $\{x,y,z,\text{roll},\text{pitch},\text{yaw}\}$ in the tool or base frame: the operator reaches a desired orientation, locks the rotational axes, and completes the translation without holding the orientation by hand; locking $x$ and $y$ gives a pure vertical insertion.

\noindent\textbf{Perching waypoints.} IL systems often add stationary cameras to an eye-in-hand camera because the egocentric view loses the goal during parts of the task~\cite{barreiros2025careful,sundaresan2024s,khazatsky2024droid,ingelhag2024robotic}; stationary cameras tie the policy to a workstation. A perching waypoint is a pre-set pose the operator returns to when useful: it brings the scene back into the wrist camera and splits a long task into shorter segments with a common start.

\noindent\textbf{Anchors.} An anchor is a zero-motion primitive: the press records the current end-effector pose, and a later primitive executes relative to it. In the mug-cleanup task of Sec.~\ref{sec:sim} the operator presses ``remember this pose''.
A constant offset measured from the demonstrations turns that pose into the drawer frame in which the carry routine places the mug. Anchors are how a learned policy can dynamically integrate information into other TAPs.

\noindent\textbf{Routines.} A robot can rotate a wrist joint continuously, flip a gripper rapidly, or execute a fixed sequence exactly; the operator's body cannot, and video games map such actions to a single button. The unscrewing routine of Sec.~\ref{sec:real} is \emph{close $\to$ rotate $\to$ open $\to$ rotate back $\to$ close $\to$ rotate $\to$ lift}, parameterised by the repetitions and the rotation per step. Routines are open-loop in this paper; a closed-loop controller or a sub-policy can be triggered the same way.

\noindent\textbf{Triggering during teleoperation.} We build on the Quest-based teleoperation of~\cite{welle2024quest2ros,van2024puppeteer,ingelhag2025real}, which tracks and relays the controllers to a Cartesian controller with adjustable gain. \taps{} are triggered by \emph{speech} (a button combination starts recording, the operator names one of the predefined commands, Whisper-tiny~\cite{radford2023robust} transcribes it and the closest command by Levenshtein distance executes, with vibration patterns confirming start, end and rejection), through an \emph{AR menu} over the controller in the headset, or by \emph{haptic buttons}: experienced operators, who do not wear the headset over the eyes, trigger the same menu blind, with vibration patterns telling them which menu is active; with the task's \taps{} pre-selected, a primitive is one press away. Every press is logged with the demonstration as a channel per primitive.

\subsection{The primitive channel} \label{sec:channel}

Let a demonstration be a sequence of observations $o_t$ and robot actions $x_t \in \mathbb{R}^{7}$ (Cartesian delta and gripper). With $K$ primitives available, we append a channel $p_t \in \{-1,+1\}^{K}$ and train the policy on the augmented action:
\begin{equation}
    a_t = [\,x_t,\ p_t\,], \qquad p_{t,k} = \begin{cases} +1 & t \in \text{label window of } k \\ -1 & \text{otherwise.} \end{cases}
\end{equation}
The label window is the primitive's routine segment or the first steps of it: the policy predicts actions in chunks, and a window starting exactly at the press teaches the channel to fire from the motion that precedes the press (the gripper closing on nothing, say); starting the window a few steps into the routine keeps the positives inside it. While a routine runs, the executed motion is recorded as ordinary actions $x_t$, so a policy trained \emph{without} the channel on the same data sees the routine as demonstrated motion. Any policy that regresses an action chunk can regress $p_t$: Diffusion Policy~\cite{chi2023diffusion}, where $p_t$ is diffused with the action, and the behavioural-cloning heads of~\cite{mandlekar2021matters}, where it is an ordinary output.

At execution the channel is decoded by a threshold: primitive $k$ is triggered on a rising edge of $\hat p_{t,k} > 0$ while no primitive is active; where several primitives share a trigger (i.e., the three waypoints of the peg task) the arg-max of the channel averaged over the last few predictions chooses. A primitive may be \emph{gated} by proprioception (a carry routine fires only with the gripper closed on something); every gate used is stated where its result is reported. Algorithm~\ref{alg:exec} gives the loop: the policy keeps predicting while a primitive runs.%
\begin{algorithm}[t]
\caption{Execution with the primitive channel}
\label{alg:exec}
active $\leftarrow$ none\;
\While{task running}{
    \If{no prediction in flight}{predict $[x_{t..t+n},\, p_{t..t+n}] \leftarrow \pi(o_t)$\;}
    \If{active $=$ none \textbf{and} rising edge on $\hat p_{t,k}$ \textbf{and} gate$_k$(proprioception)}{
        active $\leftarrow k$; $\theta \leftarrow$ parameters$_k$(anchors, proprioception)\;
    }
    \eIf{active $\neq$ none}{
        execute one step of $\rho_{\text{active}}(\theta)$\;
        \If{$\rho_{\text{active}}$ finished}{active $\leftarrow$ none; clear the action queue\;}
    }{
        execute $x_t$\;
    }
    $t \leftarrow t+1$\;
}
\end{algorithm}

\section{Simulation Study} \label{sec:sim}

The simulation study gives an unbiased answer to whether demonstrations with \taps{} produce a better policy than without, since the demonstrations are part of benchmarks that predate \taps{}.

\subsection{Adding \taps{} to unassisted demonstrations} \label{sec:sim_protocol}

\noindent\textbf{Relabelling.} We start from human teleoperation datasets in robosuite~\cite{zhu2020robosuite}: the $200$ proficient-human \emph{Square} demonstrations of robomimic~\cite{mandlekar2021matters} and a seeded subset of $200$ of the $1000$ \emph{MugCleanup D1} demonstrations of MimicGen~\cite{mandlekar2023mimicgen}. A demonstration is replayed from its recorded state up to the point an operator should have activated a \tap{}, the primitive is executed from that state with observations rendered online. Once the \tap{} ends, we stitch the demonstration back into the unassisted one and keep replaying from there. The demonstration is kept only if the task still succeeds ($\SqDemos/200$ for Square, $\MgDemos/200$ for Mug). The result is what an operator would have produced with the primitive available. The unassisted demonstrations are therefore an identical control: the exact same demonstrations, only without the \tap{}.

\noindent\textbf{Policies and evaluation.} Diffusion Policy~\cite{chi2023diffusion} in the robomimic implementation: ResNet-18 with spatial softmax on the $84\times84$ eye-in-hand image with a $76$-pixel random crop, frame stack $2$, prediction horizon $16$, $400$ epochs of $100$ steps, batch $64$, DDIM with $10$ steps; conditions differ only in the action keys and the dataset (unassisted vs TAP). Every run is evaluated at its final checkpoint, with no model selection, identically in both arms. Three training seeds per condition; $500$ seeded initial states per seed, i.e.\ $1500$ episodes per condition. We report pooled success with per-seed points, Fisher's exact test on the pooled counts, and a paired sign test on identical initial states per seed. Success is the environment's own check in both arms.

\begin{figure*}[t]
    \centering
    \includegraphics[width=0.495\linewidth]{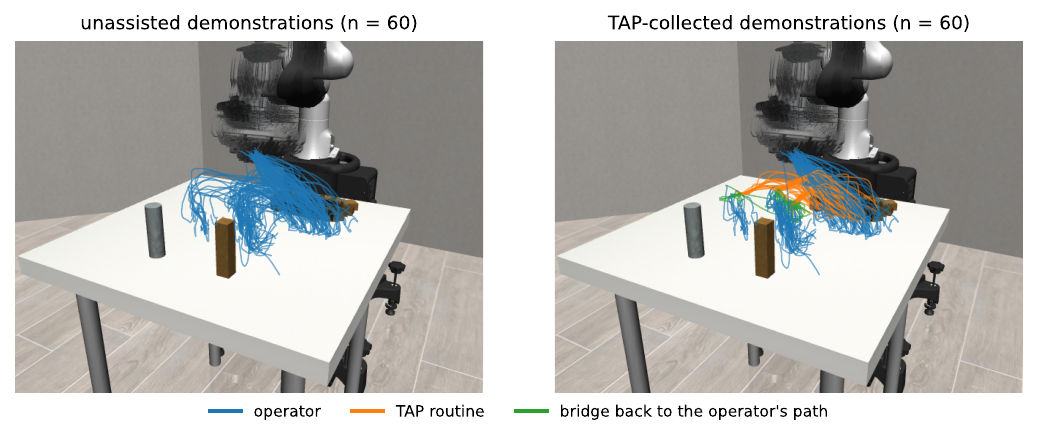}\hfill
    \includegraphics[width=0.495\linewidth]{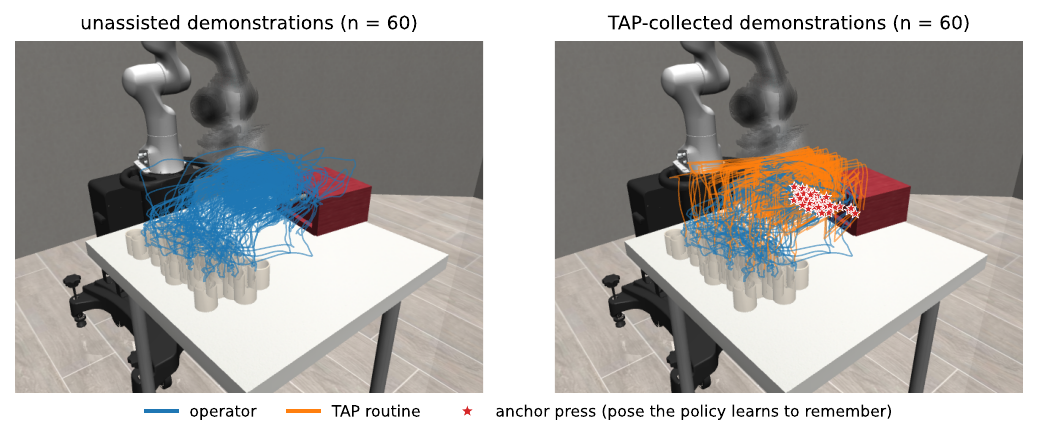}
    \vspace{-\baselineskip}
    \caption{The two simulated tasks with visualised trajectories (unassisted vs TAP). Left of each pair: unassisted demonstrations, the operator throughout. Right: the same demonstrations \tap{}-collected: the operator (blue) up to the hand-over, the primitive (orange), and on Square the stitching back to unassisted after the perch (green); on MugCleanup the star is the anchor press whose pose the policy learns to remember.}
    \label{fig:overlay}
    \vspace{-\baselineskip}
\end{figure*}

\subsection{Square: perching waypoints for wrist-cam-only policy} \label{sec:sim_square}

\noindent\textbf{Task and \tap{}.} The robot grasps a square nut and threads it onto a fixed peg ($6.75$\,mm clearance). The policy sees \emph{only} the wrist camera: no proprioception, and no fixed camera, so the peg is out of view while the nut is grasped. The \tap{} is a perching waypoint above the peg, offset to the side of the handle so that the nut hole is centred over the peg (Fig.~\ref{fig:overlay}, left). Three waypoints cover the three ways operators grasp the handle; the operator would pick the one matching the grasp. The relabelled demonstration contains the operator's grasp, the perch move, a short bridge (in green) back to the operator's trajectory, and then the operator's own recorded insertion, release, and retreat.

\noindent\textbf{The channel.} As in the real-world system, one dimension per waypoint: $+1$ on the demonstrated waypoint over a $16$-step window, $-1$ everywhere else. The rising edge of the maximum is the trigger, and the arg-max is the choice.

\noindent\textbf{Execution.}
As the policy never reads proprioceptive state, no gate before the perching waypoint is employed. Once the policy triggers the perching, we servo to the waypoint. The routine ends when the waypoint is reached, and the policy takes back control to perform the insertion. Its steps count against the same $400$-step horizon the unassisted policy is given, and both arms are scored by the identical success test.

\begin{table}[t]
\centering
\caption{Simulation, pooled over three seeds $\times$ $500$ seeded episodes. ``Stage'' is the nut lifted (Square) and the mug grasped (MugCleanup); per-seed points in Figs.~\ref{fig:square_main} and~\ref{fig:mug_main}.}
\label{tab:sim}
\footnotesize
\setlength{\tabcolsep}{3.5pt}
\begin{tabular}{llccc}
\toprule
task & demonstrations & success & stage not reached & succ.\,$|$\,stage \\
\midrule
Square & unassisted & 0.372 & 0.207 & 0.469 \\
 & \tap{} & $\mathbf{0.531}$ & 0.225 & 0.685 \\
\midrule
MugCleanup & unassisted & 0.203 & 0.509 & 0.412 \\
 & \tap{} & $\mathbf{0.323}$ & 0.534 & 0.692 \\
\bottomrule
\end{tabular}

\vspace{-\baselineskip}
\end{table}

\begin{figure}[t]
    \centering
    \includegraphics[width=0.9\linewidth]{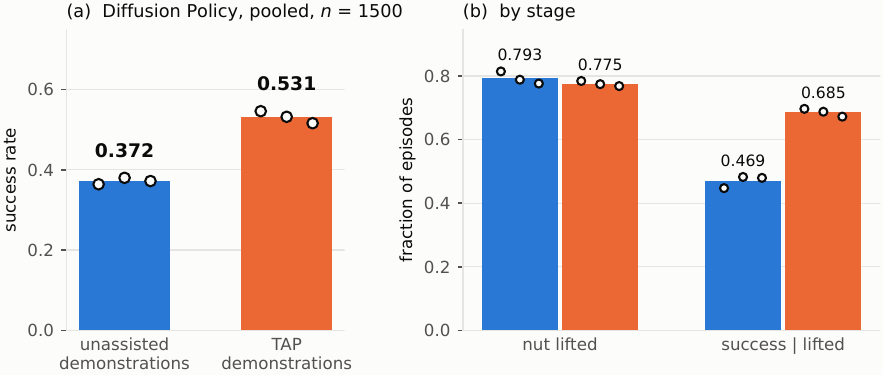}
    \caption{Square, Diffusion Policy: pooled success (a) and the same result by stage (b), both with the three training seeds as points. The lift is unchanged between the arms; only converting a lift into a success is changed.}
    \label{fig:square_main}
    \vspace{-\baselineskip}
\end{figure}

\noindent\textbf{Results.} Table~\ref{tab:sim} and Fig.~\ref{fig:square_main}: \tap{}-collected demonstrations raise pooled success from $\SqUnassisted$ to $\SqTap$ ($p = \SqP$), and \emph{all three} paired comparisons on identical initial states are significant ($\SqPairedWL$ wins, $p = \SqPairedP$). The gap is entirely downstream of the grasp: the fraction of episodes in which the nut is never lifted is unchanged ($\SqStageUn$ vs.\ $\SqStageTap$, $p = \SqPStage$), while success given a lift rises from $\SqGivenUn$ to $\SqGivenTap$ ($p = \SqPGiven$). The channel is not perfect: $\SqFiredNoLift$ of the $\SqEpisodes$ \tap{} episodes fire the routine before the nut is held and $\SqWrongPerch$ choose a waypoint that does not match the grasp; in the $\SqGoodFires$ episodes where the routine fires on a held nut with the matching waypoint, success is $\SqGoodFireSucc$.

\subsection{MugCleanup: a learned anchor used in a routine} \label{sec:sim_mug}

\noindent\textbf{Task and \taps{}.} The robot must open a drawer, place a mug upright inside, and shut the drawer; the cabinet is randomised over $20\times10$\,cm and $\pm30^{\circ}$, the mug over $40\times15$\,cm and a full turn. The policy uses the wrist camera plus proprioception, no object state; a wrist-only policy never grasps the mug ($0.005$ success), so this task is run in the regime the real system uses, and both arms see exactly the same observation. Two \taps{} are demonstrated. The \emph{anchor} press is labelled over $8$ steps around the hook pose, the last step before the drawer moves, when the open gripper sits behind the handle; nothing else changes and the operator continues. The \emph{carry} press is labelled over the first $16$ steps of the routine; from the step the mug is grasped and $3$\,cm up, the demonstration \emph{is} the carry routine: lift, traverse above the tray in the now anchored drawer frame, descend until the motion stalls, release, push the drawer shut (Fig.~\ref{fig:overlay}, right).

\noindent\textbf{Execution.} On a rising edge of the anchor channel (positive for two consecutive steps) the current end-effector pose is stored; the drawer frame follows by composition with one constant, the hook pose in the drawer frame, measured from the demonstrations. On a rising edge of the carry channel, gated on the gripper opening being in $[0.4, 7]$\,cm, the carry routine runs in that frame. The routine's parameters are the anchored frame, the current pose and gripper opening, and demonstration-derived constants (release spot, tray height, in-hand offset of the mug per gripper-opening bin). Both arms are scored with a $500$-step horizon and identical success test.

\begin{figure}[t]
    \centering
    \includegraphics[width=0.9\linewidth]{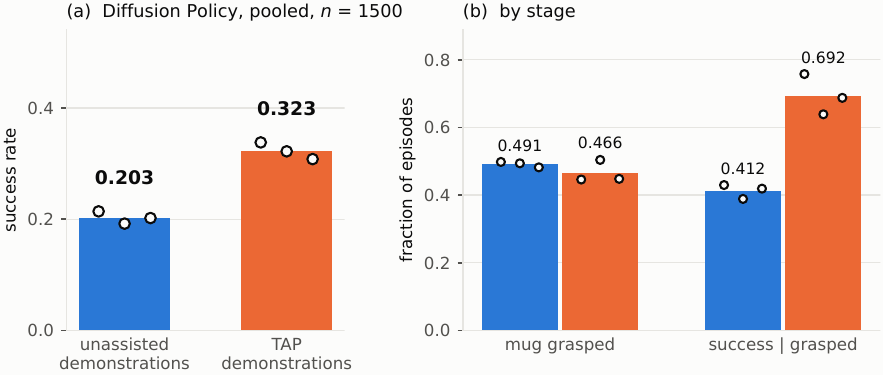}
    \caption{MugCleanup, Diffusion Policy: pooled success (a) and by stage (b), both with the three training seeds as points. Both arms grasp about equally often; converting a grasp into a success is where the \tap{} arm gains.}
    \label{fig:mug_main}
\end{figure}

\noindent\textbf{Results.} Table~\ref{tab:sim} and Fig.~\ref{fig:mug_main}: success rises from $\MgUnassisted$ to $\MgTap$ ($p = \MgP$), all three paired comparisons significant ($\MgPairedWL$ wins, $p = \MgPairedP$). Success given a grasp rises from $\MgGivenUn$ to $\MgGivenTap$ ($p = \MgPGiven$). The anchor is learned reliably: it is triggered in $\MgLatchPressed$ of $\MgEpisodes$ episodes, with a median error of $\MgLatchErrCm$\,cm and $\MgLatchErrYaw^{\circ}$ against the true drawer frame (95th percentile $\MgLatchErrCmHi$\,cm, $\MgLatchErrYawHi^{\circ}$). The carry is triggered in $\MgCarryPressed$ episodes, $\MgCarryHeld$ of them with the mug held, where success is $\MgCarryHeldSucc$. The remaining $\MgCarryEmpty$ triggers pass the gripper gate with nothing held (a gripper closed on air satisfies a gate on the opening alone), and the routine's $\approx 290$ steps then consume the episode; none of them succeeds. Both arms grasp about equally often (never grasped $\MgStageTap$ vs $\MgStageUn$, $p=\MgPStage$). The gate itself adds only $\MgGateGain$ absolute (success $\MgNogate$ with it opened wide, $p = \MgNogateP$), as it mostly guards episodes that are lost anyway: one success among the $\MgGateBlocked$ episodes in which it blocked a trigger.

\subsection{Data effect or trigger effect?} \label{sec:sim_decouple}
In order to disentangle whether the improvement comes from triggering the \tap{} or from the data itself, we score the policy trained on data \emph{without} the channel. This policy sees the routine's motion as demonstrated motion and runs alone at execution, like the unassisted policy.
The results are task-dependent (Table~\ref{tab:decouple}). On MugCleanup the no-channel policy reaches $\MgTapbase$: above the unassisted policy ($p = \MgPTapbaseVsUn$) but well below the policy with the channel ($p = \MgPTapbaseVsTap$); about a third of the gain is in the data, the regular carry being easier to imitate than the operators' own transport, and two thirds come from the learned hand-over. On Square the \tap{} data is actually slightly harmful for the policy: the no-channel policy reaches $\SqTapbase$ against $\SqUnassisted$ for its own matched control ($p = \SqPTapbaseVsUn$), while the same demonstrations with the channel reach $\SqTap$ ($p = \SqPTapbaseVsTap$ against the no-channel arm). The reason is structural: the perch routine is $\SqPctRoutine\%$ of the timesteps of a \tap{} demonstration, and the policy must imitate it from the wrist image alone, without the end-effector pose the routine itself servos on. With wrist-cam only and no features to guide, the no-channel policy fails.

\begin{table}[t]
\centering
\caption{Data effect vs.\ trigger effect, Diffusion Policy. Pooled success over three seeds $\times$ $500$ episodes. ``no channel'': the \tap{} demonstrations trained without the primitive channel, policy alone at execution.}
\label{tab:decouple}
\footnotesize
\setlength{\tabcolsep}{3pt}
\begin{tabular}{lccc}
\toprule
 & unassisted & \tap{} data, no channel & \tap{} data $+$ channel \\
\midrule
Square & $\SqUnassisted$ & $\SqTapbase$ & $\mathbf{\SqTap}$ \\
MugCleanup & $\MgUnassisted$ & $\MgTapbase$ & $\mathbf{\MgTap}$ \\
\bottomrule
\end{tabular}
\vspace{-\baselineskip}
\vspace{-\baselineskip}
\end{table}

\subsection{TAPs across policy classes} \label{sec:sim_baselines}

The primitive channel is an extra regression target, so any policy class whose loss can treat it as one can implement TAPs straightforwardly. On Square we train two robomimic behavioural-cloning baselines~\cite{mandlekar2021matters} on the same two datasets with the same image encoder, epochs, batch size, seeds and evaluation: a feed-forward MLP (BC) and a causal-Transformer BC, both with deterministic mean-squared-error heads, spanning MLP, Transformer and diffusion architectures together with Diffusion Policy itself. The channel is decoded by the same threshold rule as for Diffusion Policy.

\noindent\textbf{Result.} Every class improves on both tasks (Table~\ref{tab:baselines}). On Square, Diffusion Policy rises $\SqUnassisted$ to $\mathbf{\SqTap}$, BC $\SqBcUn$ to $\mathbf{\SqBcTap}$ ($p = \SqBcP$) and BC-Transformer $\SqBcTransformerUn$ to $\mathbf{\SqBcTransformerTap}$ ($p = \SqBcTransformerP$). On MugCleanup, Diffusion Policy rises $\MgUnassisted$ to $\mathbf{\MgTap}$, BC-Transformer $\MgBcTransformerUn$ to $\mathbf{\MgBcTransformerTap}$ ($p = \MgBcTransformerP$) and BC $\MgBcUn$ to $\mathbf{\MgBcTap}$ ($p = \MgBcP$), the latter from a floor so low that the counts matter more than the ratio: $37$ of $1500$ episodes against $111$ of $1500$. The channel therefore carries across an MLP, a causal Transformer and a diffusion head, in both observation regimes.

\begin{table}[t]
\centering
\caption{TAPs across policy classes, pooled success over three seeds $\times$ $500$ episodes. Within a task all rows share the demonstrations, split, seeds, observation, and budget.}
\label{tab:baselines}
\footnotesize
\setlength{\tabcolsep}{3pt}
\begin{tabular}{llcc}
\toprule
task & policy & unassisted & \tap{} $+$ channel \\
\midrule
Square & Diffusion Policy & $\SqUnassisted$ & $\mathbf{\SqTap}$ \\
(wrist only) & BC & $\SqBcUn$ & $\mathbf{\SqBcTap}$ \\
 & BC-Transformer & $\SqBcTransformerUn$ & $\mathbf{\SqBcTransformerTap}$ \\
\midrule
MugCleanup & Diffusion Policy & $\MgUnassisted$ & $\mathbf{\MgTap}$ \\
(wrist $+$ proprio.) & BC & $\MgBcUn$ & $\mathbf{\MgBcTap}$ \\
 & BC-Transformer & $\MgBcTransformerUn$ & $\mathbf{\MgBcTransformerTap}$ \\
\bottomrule
\end{tabular}
\vspace{-\baselineskip}
\vspace{-\baselineskip}
\end{table}

\section{Real-Robot Validation} \label{sec:real}

The real-robot experiments validate the full system end to end: demonstrations collected through the interface of Sec.~\ref{sec:method} with the operator triggering \taps{}, a Hybrid-Diffusion policy that learns to trigger the same \taps{}, and rollouts on novel initial conditions. One difference to Sec.~\ref{sec:sim} determines what they can show. In simulation we \emph{adapted} unassisted demonstrations, so the unassisted control exists. On the real robot every demonstration was collected \emph{with} \taps{}, and both policies below are trained on that data: whatever the primitives did for the operator is already in the demonstrations of both arms. The comparison therefore isolates only the learned trigger, i.e.\ it is the real-world counterpart of the ``no channel'' control of Sec.~\ref{sec:sim_decouple}. The expected outcome is parity wherever a primitive mainly helps the operator, and a difference only where the routine does something the policy cannot reproduce from its observations.

\noindent\textbf{Tasks} (Fig.~\ref{fig:taptypes}). \emph{Vial aspiration}: a syringe-equipped arm draws liquid from a vial mounted on a sphere that permits arbitrary orientation, up to the point where no liquid would flow; aligning the needle with the vial is the hard part, so the operator locks all rotational axes once the orientation is reached (axis lock). \emph{Open-container liquid transfer}: with the same syringe, draw from the right container and deposit into the left; the operator starts from a perching waypoint to draw and triggers a second waypoint that brings the target container into clear view to deposit (perching waypoints). \emph{Container unscrewing}: a second arm, not controlled by the policy, holds a container with a threaded cap; the operator positions the gripper over the cap and triggers the unscrewing routine, which grasps the cap, unscrews it and lifts it off, after which the cap is placed on the table (routine).

\noindent\textbf{Setup.} The robot is a ViperX 300\,S arm with a single eye-in-hand Intel RealSense D405 camera (Fig.~\ref{fig:taptypes} shows its view); there is no stationary camera. Demonstrations are collected with the Quest-based interface of Sec.~\ref{sec:method}; the operator triggers the \tap{} of each task where useful, and the motion the robot executes during a primitive is recorded as ordinary actions. Observations are the eye-in-hand RGB image and the 6D end-effector pose. The policy predicts an action chunk together with the \tap{} channel, executed at $10$\,Hz with a new prediction every $3$--$4$ steps; a triggered primitive takes over until it terminates while the policy keeps predicting (Sec.~\ref{sec:channel}), and if the policy indicates two primitives at the same step, the one ranked higher in the task's \tap{} library executes. The baseline Diffusion Policy (\emph{DP}) is trained on the same demonstrations without the channel: it sees each primitive's motion as demonstrated actions and must reproduce it from its observations, while \emph{Ours} is \method{} on the same data, triggering the primitives it was given. For each task we sample seven initial configurations with the same seed for both policies and execute each three times, $21$ rollouts per policy and task; a rollout counts as a success only when the task is completed. With shared demonstrations and this sample size the experiment is a validation: we read it for degradation, and for the direction of any difference, not for a significance test.

\begin{table}[t]
\centering
\caption{Real-robot results, $21$ rollouts per policy and task ($7$ configurations $\times$ $3$).
}
\label{tab:real}
\footnotesize
\setlength{\tabcolsep}{4pt}
\begin{tabular}{llcc}
\toprule
task (\tap{}) & policy & success & fail \\
\midrule
vial aspiration & Ours & $13$ ($62\%$) & $8$ \\
(axis lock) & DP & $12$ ($57\%$) & $9$ \\
\midrule
liquid transfer & Ours & $15$ ($71\%$) & $6$ \\
(perching waypoints) & DP & $13$ ($62\%$) & $8$ \\
\midrule
unscrewing & Ours & $14$ ($67\%$) & $7$ \\
(routine) & DP & $8$ ($38\%$) & $13$ \\
\bottomrule
\end{tabular}
\vspace{-\baselineskip}
\vspace{-\baselineskip}
\end{table}

\noindent\textbf{Results} (Table~\ref{tab:real}). In no task does the \tap{}-triggering policy perform worse than its counterpart. On vial aspiration and liquid transfer the two perform alike, with the hybrid policy ahead by one and two rollouts. This is the expected outcome: the axis lock and the perching waypoints made the demonstrations clean for \emph{both} policies, and a policy trained on clean demonstrations can learn to hold an orientation, or to return to a vantage point, without an explicit trigger. The waypoints still structure the demonstration, which we expect to matter more on longer-horizon tasks; the simulation study of Sec.~\ref{sec:sim_square}, which has an unassisted control, is where the effect of the perching waypoint on the policy is measured. Unscrewing reveals the channel's upside: the baseline removes the cap in $8$ of $21$ rollouts and the hybrid policy in $14$, and the baseline succeeds in all three repetitions of none of the seven configurations, whereas the hybrid policy does so in three. Inspecting the baseline failures, it turns the cap once or twice and then lifts while the cap is still threaded: a closed cap and a cap turned once look nearly identical in the images, so the policy faces a multimodal choice between turning again and lifting. The routine turns a fixed number of times, so the hybrid policy removed the cap in every rollout in which it triggered the routine correctly; its failures are triggers from a position from which the open-loop routine cannot grasp the cap. Videos of all rollouts are on the project page.

\section{Deployment: Policy-Triggered Routines for Clutch Assembly PoC} \label{sec:deploy}

\begin{figure}[t]
    \centering
    \includegraphics[width=0.80\linewidth]{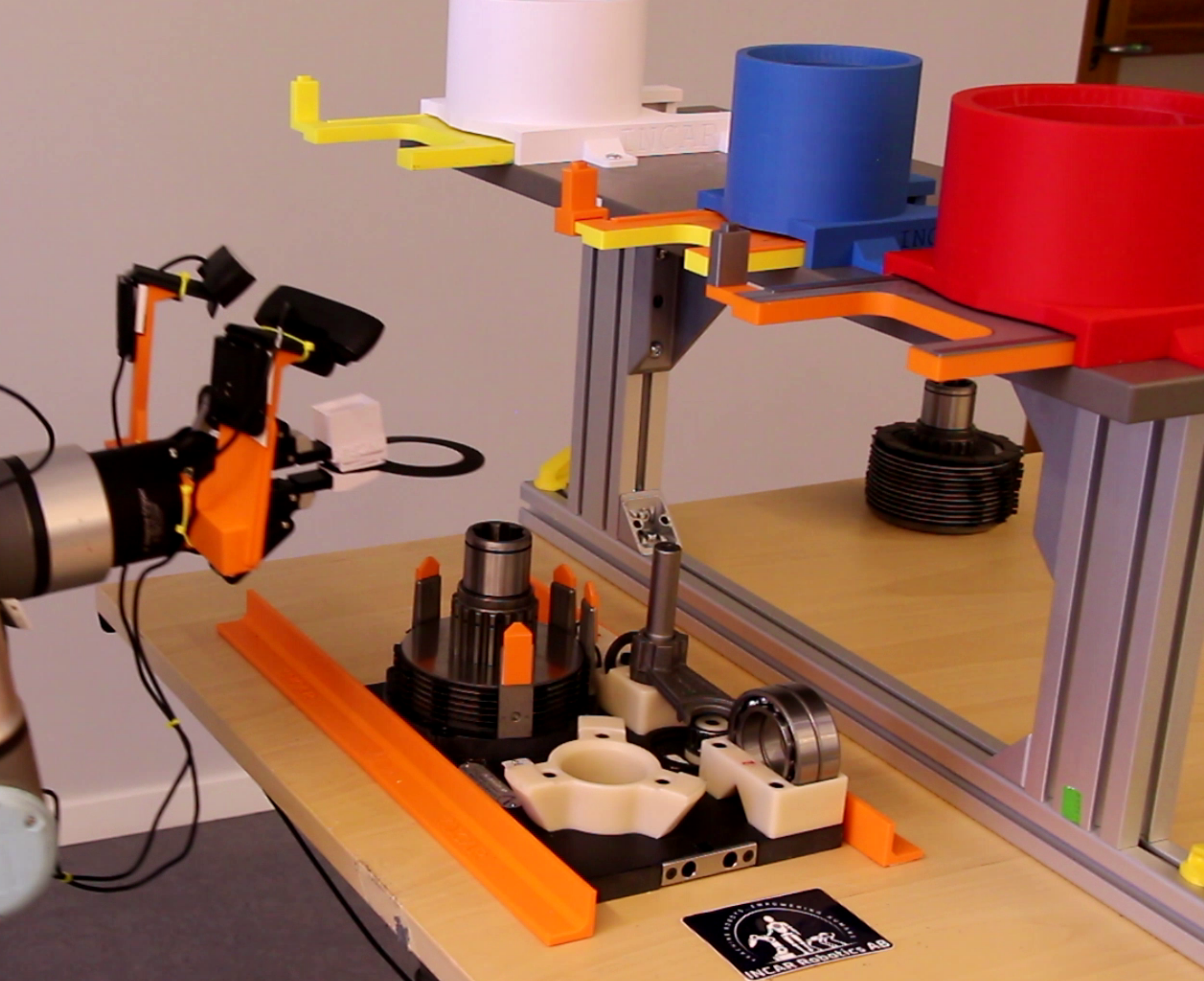}
    \caption{Clutch assembly proof-of-concept. A UR16e with a compliant two-finger gripper and two eye-in-hand cameras. The three magazines (top) hold the disc types. A routine retrieves a disc and brings it to a perch pose over the tray, where the shaft is mounted and the stack accumulates; the policy then performs the insertion and retreats, and the next routine is triggered.}
    \label{fig:poc}
    \vspace{-\baselineskip}
\end{figure}

We applied the same decomposition to an industrial clutch assembly task as a proof-of-concept: stacking the discs onto the clutch shaft of a multi-plate clutch. One clutch assembly is $35$ discs of three types (a spring disc that can face up or down, a friction disc, and a steel disc), stacked onto a shaft whose rotation is not fixed, which is mounted on a tray whose position is only roughly known. The clutch assembly is a contact-rich manipulation task. A UR16e retrieves the discs from three magazines, which an operator loads in random orientations, so the teeth are always in different orientations.
The robot is equipped with a compliant two-finger gripper and a left and a right eye-in-hand camera; no scene camera is used (Fig.~\ref{fig:poc}). Four distinct routines can be triggered to retrieve the discs, the spring disc having a normal and a flipped configuration. Each routine retrieves its disc from the magazine and brings it to a perch point over the workspace from which the tray is in view; the policy then performs the insertion and retreats, and the next routine is triggered. The routines run in a fixed order, since the assembly sequence is known and must be followed precisely.
The demonstrations were collected with the routines triggered by the operator, who teleoperated only the insertion of each disc: $15$ complete assemblies, i.e.\ $525$ demonstrated insertions. The policy's best runs assembled two complete clutches, $70$ consecutive insertions, without an interruption (video on the project page). The pattern is the one the simulation study measures: the routines make every segment except the insertion deterministic, the demonstrations are spent where perception is needed, and the \taps{} that structure the demonstration are what teach the policy when to hand over.

\section{Limitations and Conclusion}

\noindent\textbf{Limitations.} The routines in this paper are open-loop, and the policy may trigger one from a state the demonstrations do not cover: a perch fired before the nut is held, a carry fired on an empty gripper. $\SqMisfire\%$ and $\MgMisfire\%$ of the \tap{} episodes misfire on the two simulated tasks; learning a better trigger, or a routine that checks its own precondition, are potential remedies. The channel is also harder to incorporate in a Gaussian-mixture head, because a near-binary dimension drives its mixture variance to the floor and then dominates the joint likelihood, so the shared trunk is optimised almost entirely for the channel. The primitives are hand-authored per task. The real-robot experiments are small ($n=21$ per arm), control a single arm, and they share their demonstrations between the two policies, so they validate the system rather than measure the effect of \taps{} on the operator. We have not measured operator workload.

\noindent\textbf{Conclusion.} We introduced Teleoperation Augmentation Primitives (TAPs), a way for the operator to hand parts of a demonstration to the robot, and a primitive channel through which any action-chunking policy learns to do the same. In simulation, with the same demonstrations and an unassisted control, \tap{}-collected demonstrations produced better policies on the Square and MugCleanup tasks, with the gain entirely via the TAP; the control without the channel isolates what the learned trigger adds, and the channel transfers across policy classes. On a real robot, primitives that help the operator do not hurt the policy, and a routine takes over the motion a visuomotor policy cannot resolve from images alone. The same decomposition carried an industrial assembly task. We see \taps{} as a practical way to facilitate hybrid imitation learning approaches and focus the learning on the part that needs it.

\balance
\section*{Acknowledgements}
This work was supported by the Wallenberg Launch Pad (WALP) of the Knut and Alice Wallenberg Foundation.

The authors used Anthropic's Claude as an assistant for software development in the simulation setting. It contributed to the code of the simulation study of Sec.~\ref{sec:sim}, which relabels the
demonstrations, trains and evaluates the policies, and produces its tables and figures. It also corrected text and formulations in all sections. Every reported number is computed by code from recorded rollouts.

\bibliographystyle{ieeetr}
\bibliography{references}

\end{document}